\documentclass[11pt,a4paper]{article}
\usepackage[hyperref]{acl2020}
\usepackage{times}
\usepackage{latexsym}

\usepackage{amsmath}
\usepackage{graphicx, multirow}
\usepackage{microtype}

\aclfinalcopy 

\title{Variational Transformers for Diverse Response Generation}

\author{Zhaojiang Lin$^\dagger$, Genta Indra Winata, Peng Xu, Zihan Liu, Pascale Fung\\
Center for Artificial Intelligence Research (CAiRE)\\
  Department of Electronic and Computer Engineering\\
  The Hong Kong University of Science and Technology, Clear Water Bay, Hong Kong\\
  \texttt{$^\dagger$zlinao@connect.ust.hk}}

\date{}

\begin{document}
\maketitle
\begin{abstract}
Despite the great promise of Transformers in many sequence modeling tasks (e.g., machine translation), their deterministic nature hinders them from generalizing to high entropy tasks such as dialogue response generation. Previous work proposes to capture the variability of dialogue responses with a recurrent neural network (RNN)-based conditional variational autoencoder (CVAE). However, the autoregressive computation of the RNN limits the training efficiency. Therefore, we propose the Variational Transformer (VT), a variational self-attentive feed-forward sequence model. The VT combines the parallelizability and global receptive field of the Transformer with the variational nature of the CVAE by incorporating stochastic latent variables into Transformers. We explore two types of the VT: 1) modeling the discourse-level diversity with a global latent variable; and 2) augmenting the Transformer decoder with a sequence of fine-grained latent variables. Then, the proposed models are evaluated on three conversational datasets with both automatic metric and human evaluation. The experimental results show that our models improve standard Transformers and other baselines in terms of diversity, semantic relevance, and human judgment.

\end{abstract}

\section{Introduction}

Convolutional and fully-attentional feed-forward architectures, such as Transformers~\cite{vaswani2017attention}, have emerged as effective alternatives to RNNs~\cite{dehghani2018universal} in wide range of NLP tasks. These architectures remove the computational temporal dependency during the training and effectively address the long-standing vanishing gradients problem of recurrent models by processing all inputs simultaneously. Notably, transformers apply a fully attention strategy, where each token in the sequence is informed by other tokens via a self-attention mechanism. It acts as an effectively global receptive field across the whole sequences which absence in RNNs. Despite the powerful modeling capability of trasnformers, they often fail to model \textit{one-to-many}~\footnote{Given a similar dialogue history, there may exist many valid responses.} relation in dialogue response generation tasks~\cite{zhao2017learning} due to their deterministic nature. As a result, they generate dull and generic response \textit{(e.g., ``I am not sure")}, especially with greedy and beam search, which are widely used in other sequence modeling tasks. There have been attempts to generate diverse and informative dialogue responses by incorporating latent variable(s) into the RNN encoder-decoder architecture. In particular \citet{zhao2017learning} adapt a conditional variational autoencoder (CVAE) to capture discourse-level variations of dialogue, while~\citet{goyal2017z} and \citet{du2018variational} integrates latent variables in the hidden states of the RNN decoder. However, the inherently sequential computation of aforementioned models limit the efficiency for large scale training.





In this paper, we introduce the Variational Transformer (VT)~\footnote{The source code is available in \url{https://github.com/zlinao/Variational-Transformer}} a variational self-attentive feed-forward sequence model to address the aforementioned issues. The VT combine the parallelizability and global receptive field of the transformer with the variational nature of CVAE by incorporating stochastic latent variables into transformers. We explore two types of VT: 1) Global Variational Transformer (GVT), and 2) Sequential Variational Transformer. The GVT is the extension of CVAE in \citet{zhao2017learning}, which modeling the discourse-level diversity with a global  latent variable, While SVT, inspired by variational autoregressive models~\cite{goyal2017z,du2018variational}, incorporates a sequence of latent variables into decoding process by using a novel variational decoder layer. 
Unlike previous approaches~\cite{zhao2017learning,goyal2017z,du2018variational}, SVT uses \textit{Non-causal Multi-head Attention}, which attend to future tokens for computing posterior latent variables instead of using an additional encoder.

The proposed VT architectures integrate stochastic latent variables into Transformers. The experimental results on a three conversation dataset demonstrate that our models can generate more informative and coherent responses. 
\section{Related work}
\subsection{Neural Conversational Models}
Conversational systems has been widely studied~\cite{weizenbaum1966eliza,wallace2009anatomy,vinyals2015neural,serban2016generative}. Compare to rule-based systems~\cite{weizenbaum1966eliza,wallace2009anatomy}, sequence-to-sequence conversation models achieve superior performance in terms of scalable training and generalization ability~\cite{vinyals2015neural}. However, it has
been pointed out that encoder-decoder models tend to generate generic and repetitive responses like \textit{``I am sorry"}~\cite{li2016diversity}. To address this issue, there have been three main lines of work. The first is adding additional information (e.g., persona) as input to guild model generate more informative responses~\cite{li2016persona,zhang2018personalizing}. The second modifies the learning objective to promote more diverse generation~\cite{li2016diversity}, and the third integrates stochastic latent variables into Seq2Seq models by using the CVAE framework \cite{serban2017hierarchical,zhao2017learning}. Our work comes within this third line introducing a novel model, the Variational Transformer, to improve dialogue response generation.

\subsection{Conditional Variational Autoencoders}
Many works have attempted to combine CVAEs with encoder-decoder architectures for sequence generation tasks. \citet{zhang2016variational} propose a variational encoder-decoder model for neural machine translation, while \citet{li2017deep} apply variational recurrent neural networks (VRNN)~\cite{chung2015recurrent} for text summarization. \citet{zhao2017learning} and \citet{zhou2018mojitalk} explore incorporating meta features into CVAE framework in dialogue response generation tasks. \cite{goyal2017z} and \cite{du2018variational} propose variational autoregressive decoders which enhanced by highly multi-modal latent variables to capture the high variability in dialogue responses. \citet{le2018variational} further augment variational autoregressive decoders with dynamic memory networks for improving generation quality.  We unify the previous successful ideas of CVAE, and explore the combinations of CVAE and Transformer.
\subsection{Fully Attentional Networks}
Taking advantage of the parallel-in-time structure and global receptive field, Transformers~\cite{vaswani2017attention} have recently been shown to achieve impressive results on various sequence modeling tasks. Based on this, several follow-up models have been presented. The Image Transformer~\cite{parmar2018image} has been proposed for image generation, while the MultiModel~\cite{kaiser2017one} integrates convolution, attention and sparsely-gated mixture-of-expert blocks into a single deep-learning model for simultaneously learning multiple tasks from various domains. \citet{lin2019moel} proposed a fully attentional mixture-of-expert model (MoEL) for empathetic dialogue modeling. The Universal Transformer~\cite{dehghani2018universal} incorporates the recurrent inductive bias of RNNs into the standard Transformer, and achieves better result on a wide range of algorithmic and language understanding tasks. \citet{kaiser2018fast} introduce the Latent Transformer (LT) for non-autoregressive machine translation. During training, the LT first autoencodes a target sequence into a shorter sequence discrete latent variables. Then a parallel decoder decodes the target using discrete latent variables and an input sequence.
Different from the LT~\cite{kaiser2018fast}, the VT generates continuous latent variables during the decoding process. 
\section{Preliminaries}
\subsection{Conditional Variational Autoencoder for Dialogue Generation}
The CVAE framework~\cite{sohn2015learning} represents a dyadic conversation via three random variables: the input condition $c$, including conversation context and meta features (meta features can be ignored when not available); a latent variable $z$; and the target response $x$. A CVAE can be efficiently trained with Stochastic Gradient Variational Bayes (SGVB) ~\cite{kingma2013auto} by maximizing the variational lower bound of $x$ given c, according to:
\begin{equation}
    p(x | c)=\int_z p(x | z, c) p(z | c) d z.
\end{equation}

The typical CVAE consists of a prior network $p_{\theta}(z | c)$, which is used to approximate $p(z | c)$, a recognition network $p_{\phi}(z | c, x)$, which is used to approximate posterior distribution $q(z | c, x)$, and a decoder $p_{\theta}(x | z, c)$, which is used to approximate $p(x | z, c)$. By assuming z follows multivariate Gaussian distribution with a diagonal co-variance matrix, the evidence lower bound (ELBO) can be written as
\begin{equation}
    \begin{aligned} \mathcal{L}_{ELBO} &= \mathcal{L}_{REC} - \mathcal{L}_{KL}\\
    &=\mathbf{E}_{q_{\phi}(z | c, x)}\left[\log p_{\theta}(x | z, c)\right] \\ &-K L\left(q_{\phi}(z | c, x) \| p_{\theta}(z | c)\right)\\ & \leq \log p(x | c), \end{aligned}
\end{equation}
where $\mathcal{L}_{REC}$ denotes the reconstruction loss and $\mathcal{L}_{KL}$ denotes the Kullback-Leibler (KL) divergence between the posterior and prior.

In dialogue generation tasks, previous works \cite{zhao2017learning,zhou2018mojitalk} apply RNN encoders (with GRU or LSTM cell) to encode dialogue contexts and responses separately. The condition $c$ is represented by the concatenation of the last hidden state of the context encoder and the meta features (e.g., topic, emotion), while the response $x$ is represented by the last hidden state of response encoder. Then the prior network $p_{\theta}(z | c)$ and the recognition network $p_{\phi}(z | c, x)$ parameterized by multi-layer perceptrons (MLPs) are applied to approximate the means and the log variances of the prior latent distribution $\mathcal{N}\left(z ; \mu', \sigma'^{2} \mathbf{I}\right)$ and posterior latent distribution $\mathcal{N}\left(z ; \mu, \sigma^{2} \mathbf{I}\right)$. With the reparameterization trick \cite{kingma2013auto}, we can obtain samples of the prior latent variable (for testing) from $\mathcal{N}\left(z ; \mu', \sigma'^{2} \mathbf{I}\right)$ and samples of the posterior latent variable (for training) from $\mathcal{N}\left(z ; \mu, \sigma^{2} \mathbf{I}\right)$. Finally, an RNN decoder use $z$ and $c$ as the initial state to predicts the response $x$.

The \textit{vanishing latent variable problem}~\cite{bowman2016generating} is a common issue in RNN-based CVAEs. That is, the powerful autoregressive RNN decoder first learns to ignore the latent variable, and decodes the response by only condition on the previous tokens. Thus the latent variable fails to encode the meaningful information, and the CVAE deteriorates to seq2seq model. To alleviate this issue, \textit{KL annealing}~\cite{bowman2016generating} and \textit{bag-of-word loss}~\cite{zhao2017learning} have been proposed, and have shown  effectiveness in various dialogue tasks ~\cite{zhao2017learning,zhou2018mojitalk}. 

\begin{figure}[t]
\centering
\includegraphics[width=0.9\columnwidth]{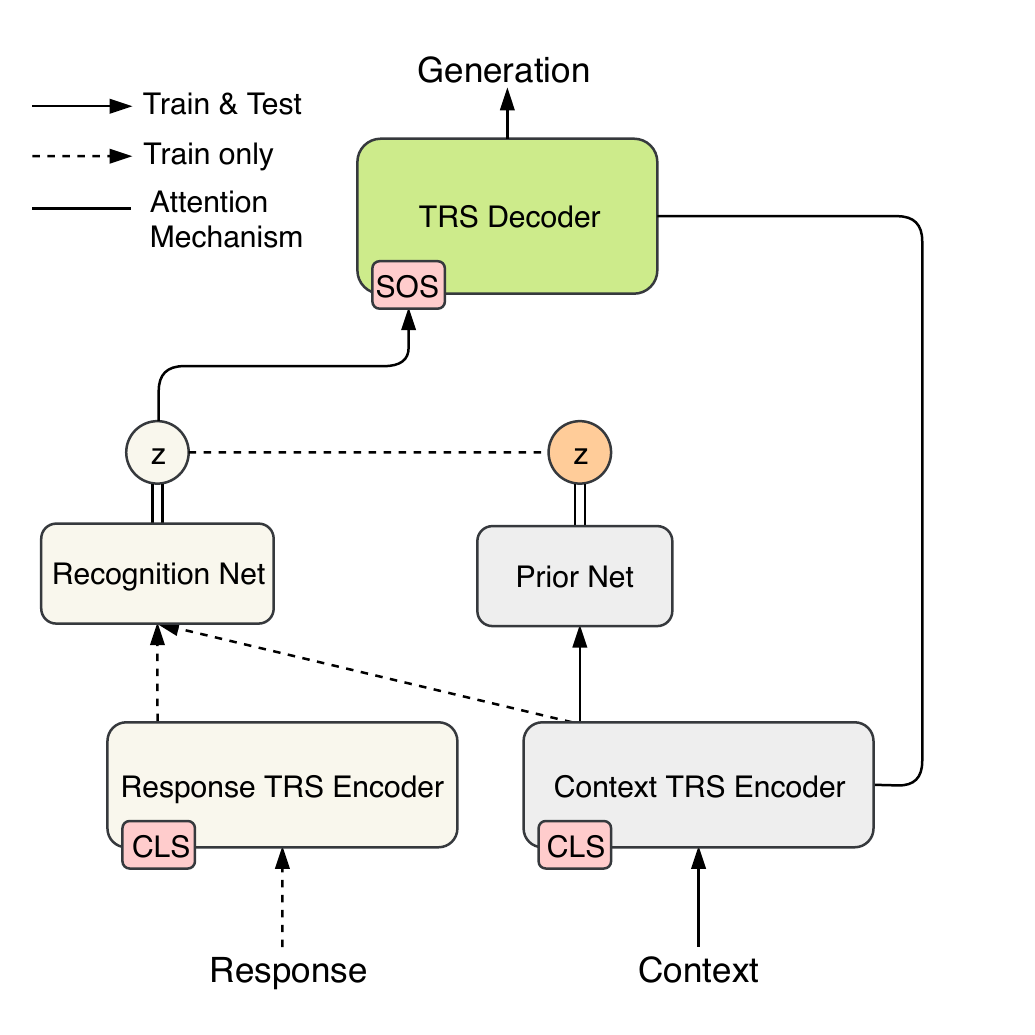} %
\caption{The Global Variational Transformer. During training, The posterior latent variable $z$ by the posterior network is passed to the decoder, while during testing, the target response is absent, and $z$ is replaced by the prior latent variable. The word embeddings, positional encoding, softmax layer and meta vectors are ignored for simplicity}
\label{vt1}
\end{figure}

\subsection{CVAE with Transformer}
The aforementioned RNN-based CVAE framework integrate the latent variable into the initial state of RNN decoder, while in transformer, it is more flexible to incorporate the latent variable embedding into the first input token of the decoder to generate the initial state. 


The overall architecture of GVT is depicted in Figure \ref{vt1}. Different from RNNs, the Transformer encoder maps an input sequence of symbol representations to a sequence of contextualized representations~\cite{vaswani2017attention}. In order to get fixed dimension representations of the response and context, we add a special token $CLS$ at the beginning of the input sequence as in BERT~\cite{devlin2018bert}, to compute the weighted sum of the output representations via self-attention. Thus the output representation of the token $CLS$ is considered as the representation of the whole sequence. Then we introduce a recognition network and a prior network to compute the posterior latent variable and prior latent variable as in \cite{zhao2017learning,zhou2018mojitalk}. We add the latent variable sample $z$ and meta features $m$ (can be ignored when not available) into $e_{SOS}$, the embedding of the \textit{start-of-sequence} token $SOS$:
\begin{equation}
   e'_{SOS} = z + m + e_{SOS} 
   \label{sos}.
\end{equation}

Finally, the transformer decoder decodes the response $x$ sequentially while attending to the new embedding $e'_{SOS}$ of token $SOS$ with latent information. 

This design enhances the CVAE framework with the global receptive field, and each position of the GVT can directly access the latent information via the multi-head self-attention mechanism. However, we still observe that the GVT suffers the \textit{vanishing latent variable problem} as RNN-based CVAE because the decoder can bypass the latent information by paying less attention to the $SOS$ token. Hence, we apply the \textit{KL annealing}, and \textit{bag-of-word} auxiliary loss $\mathcal{L}_{bow}$ as in~\cite{zhao2017learning,zhou2018mojitalk} to preserve the useful information of the latent variable. Therefore, the learning objective of the GVT is defined as follows:
\begin{equation}
    \mathcal{L} = \mathcal{L}_{ELBO} + \mathcal{L}_{bow}.
\end{equation}

\section{Sequential Variational Transformer}

In order to augment the capacity of the latent variable with multi-modal distributions and to better utilize the latent information, we further explore incorporating a sequence of latent variables in decoding process. We introduce Sequential Variational Transformer (SVT) with a novel variational decoder layer which generate latent variables for each position: $z=\left(z_{1}, \dots, z_{T}\right)$. Similar to~\citet{goyal2017z}, we interpret the latent variables as a generation plan for the future sequence.
Unlike previous CVAE models which use an extra encoder to encode the response separately~\cite{zhao2017learning,zhou2018mojitalk} or use a backward RNN to encode the future sequence for each time step~\cite{goyal2017z,du2018variational}, SVT uses a \textit{Non-causal Multi-head Attention} which leaks the future information to the recognition network for computing the posterior latent variables.

\begin{figure}[t]
\centering
\includegraphics[width=0.9\columnwidth]{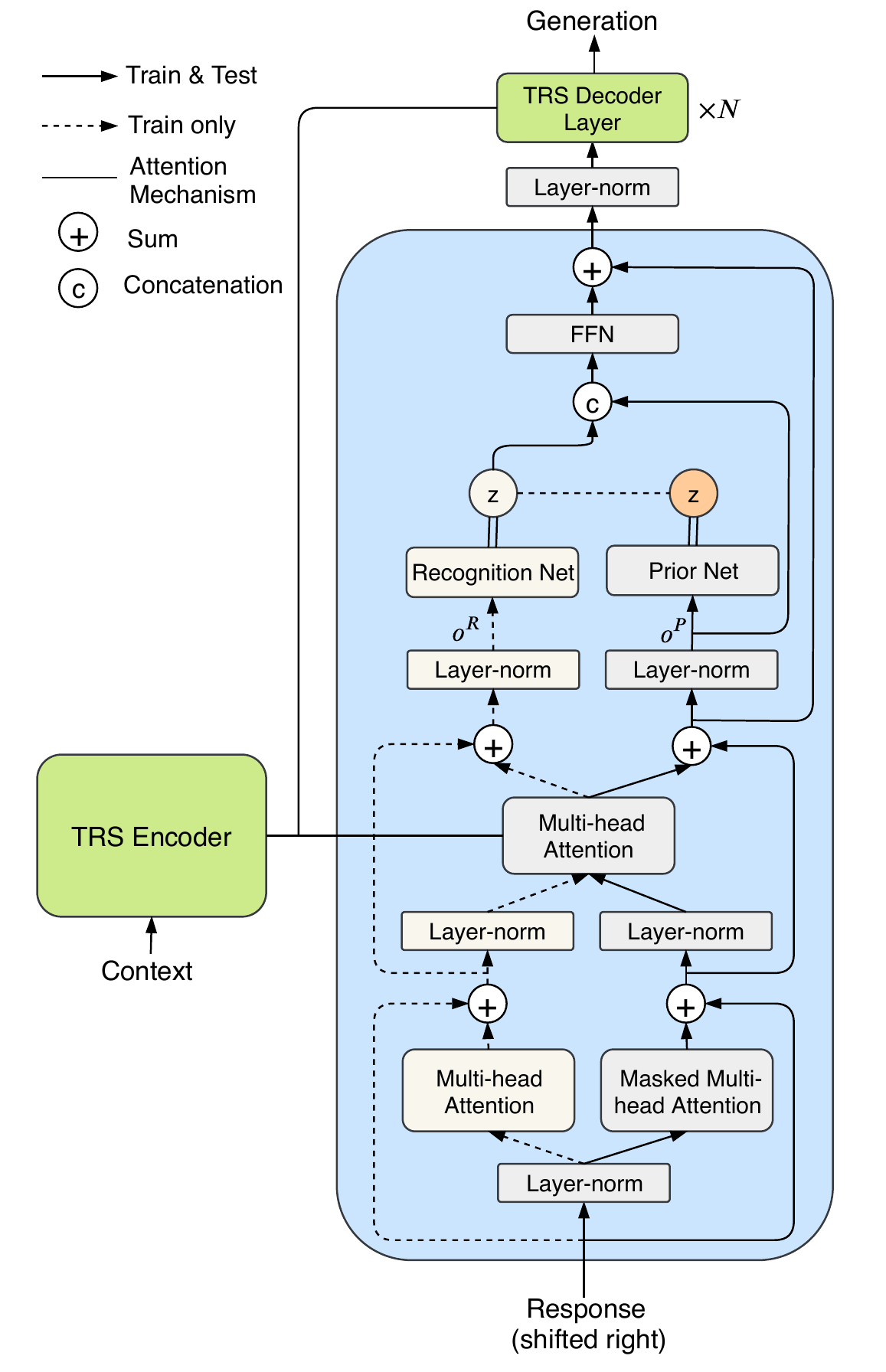} %
\caption{The Sequential Variational Transformer. During training, The posterior latent variables z by the posterior network are passed to the decoder, while during testing, the target response is absent, and z is replaced by the prior latent variables z. The word embeddings, positional encoding, softmax layer and meta vectors are ignored for simplicity}
\label{vt2}
\end{figure}

As shown in Figure \ref{vt2}, the SVT shares the same encoder as the standard Transformer~\cite{vaswani2017attention}, while its decoder consists of a variational decoder layer followed by a stack of $N$ standard Transformer decoder layers. The variational decoder layer has two paths for computing the posterior latent variable and prior latent variable respectively. We denote them as \textit{Posterior Path} and \textit{Prior Path}.

\subsection{Prior Path}
The \textit{Prior Path} (solid line in Figure \ref{vt2}) has a masked multi-head self-attention sub-layer which performs causal attention on the shifted response, followed by a multi-head self-attention sub-layer which performs encoder-decoder multi-head attention on the context encoder. The last sub-layer is composed of a MLP prior network which approximates a sequence of prior latent variable for each position, and a Position-wise Feed-Forward Network (FFN) which fuse the latent information $z$ with the observed information representation $o^P$ before the prior network (shown in Figure \ref{vt2}). Specifically, we concatenate $o^P$ with $z$ as the input to the FNN, and the FNN pass the fused representation to the next layer. Same as~\citet{vaswani2017attention}, in the variational decoder layer, each sub-layer is followed by a residual connection and layer normalization. That is, the output of each sub-layer is $LayerNorm(x + Sublayer(x))$. 

We decompose the response $x$ as $x = \left(x_1, \cdots, x_T\right)$ and the latent variable $z$ as $z=\left(z_{1}, \dots, z_{T}\right)$.  The prior model produces latent variables at each position $z_t$ by not only conditioning on the input condition $c$ (the concatenation of context and meta features), but also conditioning on the \textbf{observed response tokens} $x_{1:t-1}$. By assuming $z_t$ follows a multivariate Gaussian distribution, the prior model becomes:

\begin{equation}
p_{\theta}\left(z_{t} | c, x_{1 : t-1}\right)=\mathcal{N}\left(z_{t} ; \mu'_{t}, \sigma'_{t}\right),
\label{prior}
\end{equation}
where  $$[\mu'_{t}, \log \sigma'_{t}] = MLP(o^P).$$
\subsection{Posterior Path}
The only difference between the \textit{Posterior Path} (dash line in Figure~\ref{vt2}) and \textit{Prior Path} is that the mask is removed from the masked multi-head attention. Thus the \textbf{masked (casual) multi-head attention} become \textbf{non-casual multi-head attention}, which allows each position to attend to the subsequent positions. Then, the second multi-head attention sub-layer (shared the same weight with prior path) performs posterior attention on the encoder and passes the posterior observed information $o_R$ to the recognition network. The recognition network produces the posterior latent variable for each position $z_t$ as:
\begin{equation}
q_{\phi}\left(z_{t} | c, x\right)=\mathcal{N}\left(z_{t} ; \mu_{t}, \sigma_{t}\right),
\label{posterior}
\end{equation}
where $$[\mu_{t}, \log \sigma_{t}] = MLP(o^R).$$
During the training, the posterior path guides the learning of prior path via KL divergence constraint:
\begin{equation}
 \mathcal{L}_{KL} = \sum_{t}K L\left(q_{\phi}(z_t | c, x) \| p_{\theta}(z_t | c, x_{1:t-1})\right)
\end{equation}
In the training phase, the posterior latent variables from Equation \ref{posterior} are passed to the FFN, while in the testing phase the \textit{Posterior Path} will be blocked and the posterior latent variables will be replaced with the prior latent variables from Equation \ref{prior}. 

During the decoding process, each response token $x_t$ is generated by conditioning on observed response tokens $x_{1:t-1}$, latent variables $z_{1:t}$, and the input condition $c$. The decoding process of the SVT is:
\begin{equation}
p_{\theta}(x | z, c)=\prod_{t} p_{\theta}\left(x_{t} | z_{1 : t}, x_{1 : t-1}, c\right).
\end{equation}

\subsection{Auxiliary Loss}
As we expect the latent variables to be a generation plan for the future sequence, we inject such bias into latent variables by using an auxiliary loss: \textit{Sequential-Bag-of-Word (SBOW)} which proposed by \citet{du2018variational}. The idea of the \textit{SBOW} auxiliary objective is to sequentially predict the bag of succeeding target words $x_{t:T}$ by using latent variable $z_t$. In our case, the succeeding words prediction also leverages the observed information $c$ and $x_{1:t-1}$. Thus the auxiliary loss at each position is computed by: 
\begin{equation}
   p_{\xi}(x_{t:T}|z_t,c,x_{1:t-1}) = f_{aux}(z_t,o^P) 
\end{equation}
where $f_{aux}$ is a feed-forward neural network with the softmax output.

\subsection{Learning}
The evidence lower bound (ELBO) objective of SVT is the sum of the reconstruction loss $\mathcal{L}_{REC}(t)$ and Kullback-Leibler divergence loss $\mathcal{L}_{KL}(t)$ at each position:

\begin{equation}
    \begin{aligned}
    \mathcal{L}_{ELBO} &= \sum_{t} \mathcal{L}_{REC}(t) - \mathcal{L}_{KL}(t)\\
    &=\sum_{t}\mathbf{E}_{q_{\phi}(z_t | c, x)}\left[\log p_{\theta}(x_t | z_{1:t},x_{1:t-1}, c)\right] \\ &-K L\left(q_{\phi}(z_t | c, x) \| p_{\theta}(z_t | c, x_{1:t-1})\right).
    \end{aligned}
    \label{elbo}
\end{equation}

We regularize the ELBO learning objective with an auxiliary loss $\mathcal{L}_{sbow}$ to enhance the expressiveness of the latent variables. Therefore, the final learning objective is formulated as follows:
\begin{equation}
\mathcal{L} = \mathcal{L}_{ELBO} + \mathcal{L}_{sbow},
\end{equation}
where,
\begin{equation}
\mathcal{L}_{sbow} = \sum_{t}\mathbf{E}_{q_{\phi}(z_t | c, x)}\left[\log p_{\xi}(x_{t:T} | z_t,x_{1:t-1}, c)\right].
\end{equation}

\section{Experiments}
\subsection{Dataset}
We evaluate the proposed models on three conversationet dataset such as MojiTalk~\cite{zhou2018mojitalk}, PersonaChat~\cite{zhang2018personalizing}, Empathetic-Dialogues~\cite{rashkin2019towards}.

\paragraph{MojiTalk} dataset consists of 596,959 post and response pairs from Twitter. Each response is labeled by one emoji which indicates the response emotion. There are 64 emoji labels in total with unbalanced distribution. We use the preprocessed data and vocabulary released from ~\citet{zhou2018mojitalk} and follow the same split of train/validation/test set.

\paragraph{PersonaChat \& Empathetic-Dialogues} are one-to-one multi-turn conversation datasets. In PersonaChat (Persona), the conversations are revolve around personas which are established by four to six persona sentences. While in Empathetic-Dialogues (ED), the conversation are mostly about situation that happened to one of the speaker and another speaker is trying to understand the feeling and reply accordingly. Both datasets are about modeling social skills and the goal is to make user more engaging. Therefore, we combine the train/validation/test set of two datasets.
\subsection{Baselines}
We compare the proposed models with the following baselines:
\paragraph{Seq2Seq.} An attention-based
sequence-to-sequence model with the emoji vector as additional input as discribed in MojiTalk~\cite{zhou2018mojitalk}.
\paragraph{CVAE.} An RNN-based conditional variational autoencoder for dialogue response generation~\cite{zhou2018mojitalk}, which uses a multivariate Gaussian latent variable to model the response and concatenate it with the last hidden state of the encoder as the initial state of the decoder. KL annealing, early stopping strategy and \textit{bag-of-word} auxiliary loss are applied during the training. We use the implementation~\footnote{The implementation of CVAE baseline: \url{https://github.com/claude-zhou/MojiTalk}} released by ~\citet{zhou2018mojitalk}.

\paragraph{Transformer.} 
A transformer~\cite{vaswani2017attention} trained by using a Maximum Likelihood Estimation (MLE) objective and can be considered as the base model for both the GVT and SVT.  

\subsection{Hyper-parameters and Training Setup}
We use a 4-layer Transformer as our base model. The hidden size is set to be 300 everywhere, and the word embedding is initialized with the 300-dimensional pre-trained GloVe embeddings for both encoder and decoder. The multi-head attention sub-layers are made up of 4 attention heads each with embedding dimension 64. The size of latent variable is 300. The recognition network and the prior network are parameterized by 3-layer MLPs with 512 hidden dimension. Following the training setup of ~\citet{zhou2018mojitalk}, we first train our baseline transformer model with the MLE objective and use it to initialize its counterparts in both GVT and SVT.  Then the models are trained end-to-end by the Adam optimizer with  the initial learning rate $2\times10^{-4}$. KL annealing and early stopping strategy are applied as in~\cite{zhou2018mojitalk}.
In the test time, we use greedy decoding strategy for all models.

\begin{table*}[!ht]
\centering
\resizebox{0.99\textwidth}{!}{
\begin{tabular}{lccccccccc}
\hline \hline
\multicolumn{10}{l}{\textbf{MojiTalk}}  \\ \hline
\multicolumn{1}{c|}{\multirow{2}{*}{\textbf{Model}}} & \multicolumn{1}{c|}{\multirow{2}{*}{\textbf{PPL}}} & \multicolumn{1}{c|}{\multirow{2}{*}{\textbf{KLD}}} & \multicolumn{3}{c|}{\textbf{Diversity}}  & \multicolumn{2}{c|}{\textbf{Embeddings Similarity}}  & \multicolumn{2}{c}{\textbf{Human Evaluation}}  \\ \cline{4-10} 
\multicolumn{1}{l|}{}  & \multicolumn{1}{c|}{}  & \multicolumn{1}{c|}{}  & \multicolumn{1}{c|}{\textbf{Dist-1}} & \multicolumn{1}{c|}{\textbf{Dist-2}} & \multicolumn{1}{c|}{\textbf{Dist-3}} & \multicolumn{1}{c|}{$\textbf{EMB}_\textbf{FT}$}  & \multicolumn{1}{c|}{$\textbf{EMB}_\textbf{BERT}$}  & \multicolumn{1}{c|}{\textbf{Coherence}} & \textbf{Emotion}  \\ \hline
\multicolumn{1}{l|}{Seq2Seq}  & \multicolumn{1}{c|}{130.75}  & \multicolumn{1}{c|}{-}  & \multicolumn{1}{c|}{0.0055}  & \multicolumn{1}{c|}{0.0187}  & \multicolumn{1}{c|}{0.0347}  & \multicolumn{1}{c|}{0.738}  & \multicolumn{1}{c|}{0.594}  & \multicolumn{1}{c|}{20.67}  & 20.67  \\
\multicolumn{1}{l|}{CVAE}  & \multicolumn{1}{c|}{35.33}  & \multicolumn{1}{c|}{27.55}  & \multicolumn{1}{c|}{0.0189}  & \multicolumn{1}{c|}{0.1340}  & \multicolumn{1}{c|}{0.3640}  & \multicolumn{1}{c|}{0.751}  & \multicolumn{1}{c|}{0.613}  & \multicolumn{1}{c|}{18.33}  & 18  \\
\multicolumn{1}{l|}{Transformer}  & \multicolumn{1}{c|}{72.66}  & \multicolumn{1}{c|}{-}  & \multicolumn{1}{c|}{0.0040}  & \multicolumn{1}{c|}{0.0161}  & \multicolumn{1}{c|}{0.0324}  & \multicolumn{1}{c|}{0.741}  & \multicolumn{1}{c|}{0.596}  & \multicolumn{1}{c|}{19.67}  & 23.33  \\ \hline
\multicolumn{1}{l|}{GVT}  & \multicolumn{1}{c|}{19.71}  & \multicolumn{1}{c|}{18.15}  & \multicolumn{1}{c|}{\textbf{0.0207}} & \multicolumn{1}{c|}{\textbf{0.1524}} & \multicolumn{1}{c|}{\textbf{0.4064}} & \multicolumn{1}{c|}{0.753}  & \multicolumn{1}{c|}{0.609}  & \multicolumn{1}{c|}{23}  & 22.67  \\
\multicolumn{1}{l|}{SVT}  & \multicolumn{1}{c|}{\textbf{18.96}}  & \multicolumn{1}{c|}{32.27}  & \multicolumn{1}{c|}{0.0079}  & \multicolumn{1}{c|}{0.1053}  & \multicolumn{1}{c|}{0.3654}  & \multicolumn{1}{c|}{\textbf{0.762}}  & \multicolumn{1}{c|}{\textbf{0.619}}  & \multicolumn{1}{c|}{\textbf{26}}  & \textbf{27.67}  \\ \hline
\multicolumn{1}{l|}{Human}  & \multicolumn{1}{c|}{-}  & \multicolumn{1}{c|}{-}  & \multicolumn{1}{c|}{0.0557}  & \multicolumn{1}{c|}{0.4009}  & \multicolumn{1}{c|}{0.7697}  & \multicolumn{1}{c|}{-}  & \multicolumn{1}{c|}{-}  & \multicolumn{1}{c|}{-}  & -  \\ \hline \hline
\multicolumn{10}{l}{\textbf{Persona + ED}}  \\ \hline
\multicolumn{1}{c|}{\multirow{2}{*}{\textbf{Model}}} & \multicolumn{1}{c|}{\multirow{2}{*}{\textbf{PPL}}} & \multicolumn{1}{c|}{\multirow{2}{*}{\textbf{KLD}}} & \multicolumn{3}{c|}{\textbf{Diversity}}  & \multicolumn{2}{c|}{\textbf{Embeddings Similarity}}  & \multicolumn{2}{c}{\textbf{Human Evaluation}}  \\ \cline{4-10} 
\multicolumn{1}{c|}{}  & \multicolumn{1}{c|}{}  & \multicolumn{1}{c|}{}  & \multicolumn{1}{c|}{\textbf{Dist-1}} & \multicolumn{1}{c|}{\textbf{Dist-2}} & \multicolumn{1}{c|}{\textbf{Dist-3}} & \multicolumn{1}{c|}{\textbf{$\textbf{EMB}_\textbf{FT}$}} & \multicolumn{1}{c|}{\textbf{$\textbf{EMB}_\textbf{BERT}$}} & \multicolumn{1}{c|}{\textbf{Coherence}} & \textbf{Engagedness} \\ \hline
\multicolumn{1}{l|}{CVAE}  & \multicolumn{1}{c|}{31.32}  & \multicolumn{1}{c|}{10.01}  & \multicolumn{1}{c|}{0.0186}  & \multicolumn{1}{c|}{0.1102}  & \multicolumn{1}{c|}{0.295}  & \multicolumn{1}{c|}{\textbf{0.917}}  & \multicolumn{1}{c|}{0.666}  & \multicolumn{1}{c|}{20.67}  & 21.33  \\
\multicolumn{1}{l|}{Transformer}  & \multicolumn{1}{c|}{48.03}  & \multicolumn{1}{c|}{-}  & \multicolumn{1}{c|}{0.0058}  & \multicolumn{1}{c|}{0.0237}  & \multicolumn{1}{c|}{0.0524}  & \multicolumn{1}{c|}{0.915}  & \multicolumn{1}{c|}{0.672}  & \multicolumn{1}{c|}{24.67}  & 24.67  \\ \hline
\multicolumn{1}{l|}{GVT}  & \multicolumn{1}{c|}{18.34}  & \multicolumn{1}{c|}{19.13}  & \multicolumn{1}{c|}{0.0204}  & \multicolumn{1}{c|}{0.1406}  & \multicolumn{1}{c|}{\textbf{0.3995}}  & \multicolumn{1}{c|}{\textbf{0.917}}  & \multicolumn{1}{c|}{\textbf{0.675}}  & \multicolumn{1}{c|}{20}  & 21.33  \\
\multicolumn{1}{l|}{SVT}  & \multicolumn{1}{c|}{\textbf{17.75}}  & \multicolumn{1}{c|}{24.67}  & \multicolumn{1}{c|}{\textbf{0.0213}}  & \multicolumn{1}{c|}{\textbf{0.1521}}  & \multicolumn{1}{c|}{0.3936}  & \multicolumn{1}{c|}{0.906}  & \multicolumn{1}{c|}{0.665}  & \multicolumn{1}{c|}{\textbf{38.67}}  & \textbf{36.67}  \\ \hline
\multicolumn{1}{l|}{Human}  & \multicolumn{1}{c|}{-}  & \multicolumn{1}{c|}{-}  & \multicolumn{1}{c|}{0.0640}  & \multicolumn{1}{c|}{0.3800}  & \multicolumn{1}{c|}{0.7070}  & \multicolumn{1}{c|}{-}  & \multicolumn{1}{c|}{-}  & \multicolumn{1}{c|}{-}  & -  \\ \hline \hline
\end{tabular}
}
\caption{Results of Variational Transformer compared to baselines on automatic and human evaluations.}
\label{results}
\end{table*}

\subsection{Automatic Evaluation}

\paragraph{PPL \& KLD.} 
The  evaluation  metrics  include Perplexity (\textbf{PPL}) and
Kullback-Leibler divergence between the posterior and prior (\textbf{KLD}). A well trained model should achieve a low reconstruction and small but non-trivial KL distance~\cite{zhao2018unsupervised}.

\paragraph{Diversity.} 
To measure the generation diversity, we calculate \textbf{Dist-1}, \textbf{Dist-2}, and \textbf{Dist-3}, the ratio of the number of distinct n-grams (unigrams, bigrams, and trigrams) over the total number of n-grams. A higher distinct n-grams ratio indicates more diverse generation.

\paragraph{Embeddings Similarity.}
This metric computes the cosine similarity between the sentence embedding of a generated sequence and that of a ground-truth response. In our experiments, we introduce two different ways to represent sentence embeddings. The first is  $\textbf{EMB}_\textbf{FT}$~\cite{liu2016not} that calculates the average of word embeddings in a sentence using FastText ~\cite{mikolov2018advances} which is trained with Common Crawl and Wikipedia data. We use FastText embeddings instead of other pre-trained word embeddings because it can handle out-of-vocabulary issue. However, representing a sentence by simply taking the average of word embeddings ignores the context information. Therefore, we propose to use a pre-trained language model BERT~\cite{devlin2018bert} to compute the contextualized sentence representation. Specifically, we use a pre-trained BERT to encode a generated sentence and a ground-truth response, and average the output representation of both to obtain the sentence embeddings. We denote such contextualized sentence embedding as $\textbf{EMB}_\textbf{BERT}$.

\subsection{Human Evaluation}
In the human evaluation, we prepare multiple-choice questions for human evaluators and the answers are the generation results from the five models (Seq2Seq, CVAE, Transformer, GVT, and SVT). we  first randomly sample 100 dialogues and their corresponding responses from our models and the baselines. For each response, we assign three human annotators to select the most coherent (on topic) response to the context (multiple answers are allowed). In addition, annotators also need to choose the best response correlated to the given emoji label in Mojitalk and the most engaging response in PersonaChat and Empathetic-Dialogues. 
If there is no response that satisfies the evaluators, they can choose ``all answers are bad", which means none of the answer is chosen. We compute the rate that each model is chosen to quantify generation quality regarding to the human standard.


\section{Results}
\subsection{Quantitative Analysis}
The automatic evaluation results are shown in Table~\ref{results}. Transformer-based models have significantly lower perplexity compared to RNN-based models which indicate that the global receptive field performed by multi-head self-attention boost the modeling capacity. However, deterministic Seq2Seq and Transformer models tends to generate generic responses which leads to a low diversity score. Meanwhile incorporating a stochastic latent variable into both models (CVAE and GVT) promote more diverse generation results and boost the diversity scores such as \textbf{Dist-1}, \textbf{Dist-2}, and \textbf{Dist-3}.

\begin{table*}[!ht]
\centering
\resizebox{0.83\textwidth}{!}{
\begin{tabular}{l|l}
\hline
Context & trade must 've made you mad ? \\ \hline
Emotion & \includegraphics[height=0.35cm]{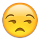} \\ \hline
\multirow{6}{*}{Responses} & \textbf{Seq2Seq:} i 'm not sure if i 'm not sure if i 'm not sure if i 'm not sure about it \\
 & \textbf{CVAE:} \textless unk\textgreater\text{ }but i don 't think it 's been on \\
 & \textbf{Transformer:} i 'm not sure i 'm not \\
 & \textbf{GVT:} i 'll pass it on , she 's mad \\
 & \textbf{SVT:} hell yeah bro . yeah \\ 
 & \textbf{Ref:} i don 't wanna talk about it \\ \hline \hline
 Context & love the smell of a good bbq ! \includegraphics[height=0.35cm]{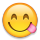} \includegraphics[height=0.35cm]{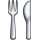} \\ \hline
Emotion & \includegraphics[height=0.35cm]{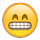} \\ \hline
\multirow{6}{*}{Responses} & \textbf{Seq2Seq:} i love it ! i love it ! \\
 & \textbf{CVAE:} aw you 're getting better \\
 & \textbf{Transformer:} i 'm glad you like it ! \\
 & \textbf{GVT:} i 'd like to know you 're very well . \\
 & \textbf{SVT:} omg what kind of smell ? thanks for sharing it with a pizza in \textless unk\textgreater \\ 
 & \textbf{Ref:} hahaha that sounds like friday to me pc \\ \hline \hline
 
 Context & \begin{tabular}[c]{@{}l@{}}good evening , how are you tonight ?\end{tabular} \\ \hline

\multirow{6}{*}{Responses}
 & \textbf{CVAE:}  hi , i am doing well . you ?\\
 & \textbf{Transformer:}  i am doing well , how are you ? \\
 & \textbf{GVT:}  i am good . just studying some people there . \\
 & \textbf{SVT:} : i am doing well , just finished eating some ice cream .\\ 
 & \textbf{Ref:} i am doing well . i am relaxing before bed . i work in the morning . \\ \hline
 
 \hline

   Context       & i cook mine at home while watching one tree hill . love that show . \\
 
 \hline

\multirow{6}{*}{Responses} 
 & \textbf{CVAE:}  i love to cook . i like to cook .\\
 & \textbf{Transformer:}  i love the outdoors . i love the outdoors .\\
 & \textbf{GVT:}it is good . . . you can make some money .\\
 & \textbf{SVT:} do you have any pets ? i have never watched any of tv. \\ 
 & \textbf{Ref:} i am looking for a new job . i hate sitting still all day \\ 
 \hline\hline
  &  User: well do you work ? do you have a degree to sustain yourself ? \\
   Context       & System:  i built models when i was a kid . now i sculpt and mold and carve. \\
          & User: nice , i specialize in computer science degree so i mostly mold 3d images.\\
 \hline

\multirow{6}{*}{Responses} 
 & \textbf{CVAE:}  i do not like it when you get to do the same \\
 & \textbf{Transformer:}  i am a teacher . i am a teacher . i am a teacher . \\
 & \textbf{GVT:} me too ! my favorite is the best baker .\\
 & \textbf{SVT:}  i love the technology . i like to play when i get older \\ 
 & \textbf{Ref:} i am looking for a new job . i hate sitting still all day \\ \hline \hline
\end{tabular}
}
\caption{Generated responses from proposed models and baseline models. The reference responses (Ref) are given.}
\label{generation}
\end{table*}


Compare to baseline models, the GVT achieves relatively lower reconstruction \textbf{PPL},  which suggests that the global latent variable contains rich latent information (e.g., topic) for response generation.  Meanwhile, the sequential latent variables of the SVT encode fine-grained latent information and further improve the reconstruction \textbf{PPL}.


On the other hand, SVT achieves the highest score in terms of two semantic relevance-oriented metrics such as $\textbf{EMB}_\textbf{FT}$ and $\textbf{EMB}_\textbf{BERT}$ in MojiTalk dataset, while in the combined dataset of Persona and ED, we observe performance drop of SVT compare to other models. This is because both Persona and ED are well designed and have lower entropy than MojiTalk which collected from Twitter. We hypothesize that the sequential latent variables have no advantage in term of similarity to single, fixed "gold response" when model low entropy response. Indeed, in open domain dialogue response generation, automatic metric is not always aligned with the human judgement~\cite{liu2016not}. In contrast, human evaluation result reported in Table \ref{results} demonstrates the generations of SVT are closer to the human standard in terms of coherence, invoked emotion and engagedness.

\subsection{Qualitative Analysis}
Table \ref{generation} compares the generation of the proposed models with baselines given the same contexts. We observe that the Seq2Seq and vanilla transformer tend to generate generic and repetitive responses (e.g., i am not sure) in MojiTalk due to their deterministic structure fail to capture the variability in dialogue response. By incorporating stochastic latent variables, the CVAE and GVT can generate more diverse responses, but their responses are sometimes digressive (e.g., example 5). Interestingly, GVT and SVT generalize the topic beyong the context which make the dialogue more engaging (e.g., example 4). In general, SVT is able to generate more coherent and informative responses.



\section{Conclusion}
This paper introduces the Variational Transformer (VT), a variational self-attentive feed-forward sequence model that combines the global receptive field of a Transformer with the variational nature of a CVAE. We propose two types of the VT:  1) the Global Variational Transformer (GVT) which incorporates a global latent variable as additional input to the transformer decoder; and 2) the Sequential Variational Transformer (SVT) which generates latent variables for each position during decoding process. Quantitative and qualitative experimental results shows that our models outperform baselines in terms of diversity, semantic relevance, and human judgment. In future work, we will utilize the pre-training language models~\cite{radford2019language} as the back-bone to strengthen the language model of the VT for better generation.

\newpage

\bibliography{acl2020}
\bibliographystyle{acl_natbib}




\end{document}